\documentclass[conference]{IEEEtran}
\IEEEoverridecommandlockouts

\usepackage{cite}
\usepackage{amsmath,amssymb,amsfonts}
\usepackage{algorithmic}
\usepackage{graphicx}
\usepackage{textcomp}
\usepackage{xcolor}
\usepackage{stfloats}

\def\BibTeX{{\rm B\kern-.05em{\sc i\kern-.025em b}\kern-.08em
    T\kern-.1667em\lower.7ex\hbox{E}\kern-.125emX}}

\begin{document}

\title{Fed-Equilibrium Framework for Topological Pareto Control in Robust and Fair Clinical Federated Learning\\
\thanks{This work was supported by the Natural Sciences and Engineering Research Council of Canada (NSERC). The authors gratefully acknowledge the Canadian Network for Observational Drug Effect Studies (CNODES) and the MITRE Corporation (SyntheticMass) for providing the synthetic datasets used in this research.}
}

\author{
\IEEEauthorblockN{Ting Xu}
\IEEEauthorblockA{\textit{Department of Biomedical Engineering} \\
\textit{University of Calgary}\\
Calgary, Canada \\
0009-0009-7461-8237}
\and
\IEEEauthorblockN{Henry Leung}
\IEEEauthorblockA{\textit{Department of Electrical and Software Engineering} \\
\textit{University of Calgary}\\
Calgary, Canada \\
0000-0002-5984-107X}
}

\maketitle

\begin{abstract}
The deployment of Federated Learning (FL) in multi-center clinical networks faces the challenge of ``knowledge dominance,'' where high-volume hubs naturally overwhelm minority community nodes, implicitly treating the distinct clinical patterns of smaller cohorts as outliers. Existing geometric defenses provide a security baseline but leave this efficiency-fairness dilemma unresolved. To bridge this gap, we propose Fed-Equilibrium, a framework that advances the paradigm from simple defense to topological equilibrium. Unlike traditional aggregators, Fed-Equilibrium implements a sequential architectural synergy. It utilizes a two-stage gradient control cascade: Stage~I (geometric quality assurance) enforces directional consistency via a cosine similarity funnel to filter malicious noise, creating a stabilized manifold; Stage~II (topological Pareto control) then actively modulates verified contributions by identifying the optimal Pareto knee point. We validated this framework on a bi-national simulation integrating Canadian (CNODES) and U.S. (SyntheticMass) registries. Experimental results demonstrate that the system simultaneously secures the network against adversarial divergence while accommodating underrepresented signals. Notably, the minority U.S. spoke (representing less than 3\% of data volume) achieved deep convergence comparable to the data-rich Canadian hub. This confirms that Fed-Equilibrium effectively counters ``knowledge dominance,'' establishing a true ``knowledge commons'' where global generalizability does not come at the cost of local clinical representation.
\end{abstract}


\section{Introduction}

Clinical registries have evolved into the foundation of modern medical research, offering real-world evidence that complements randomized controlled trials \cite{ref1, ref2}. However, the digitization of healthcare has created a fragmentation paradox: while massive volumes of high-value clinical data exist, they remain locked within institutional ``data silos'' due to stringent privacy regulations such as HIPAA and GDPR \cite{ref3}. Federated Learning (FL) has emerged as the standard solution to bridge these silos, enabling collaborative intelligence without requiring direct data exchange \cite{ref4}.

Nevertheless, in securing these collaborative networks, the research community has largely focused on defense rather than equilibrium \cite{ref5}. Prior works, such as the geometric quality assurance (GQA) framework, have successfully established a robust security baseline, effectively filtering out adversarial divergence by enforcing directional consistency via a cosine similarity funnel \cite{ref6, ref7}. Yet, security alone does not guarantee clinical fairness, especially in real-world federated networks like PCORnet \cite{ref8} or OHDSI \cite{ref9}. In these influence-asymmetric structures, large hubs (e.g., tertiary hospitals) tend to dominate and neglect the voices of smaller spokes (e.g., community clinics).

Clinical data is inherently non-IID (independent and identically distributed) \cite{ref10}. Even within a secure network protected by geometric defenses, a critical secondary challenge remains: knowledge dominance \cite{ref11}. Large tertiary centers (dominant nodes) generate high-magnitude gradients that naturally overwhelm the subtle updates from smaller community clinics (minority nodes). Standard aggregation algorithms like FedAvg suffer from Pareto inefficiency by blindly averaging these updates \cite{ref4, ref11, ref12}. As a result, even though minority data, containing valuable rare disease profiles, is successfully preserved by the security filter, its impact can be overlooked during the aggregation process. The global model inevitably converges towards the ``average'' patient, failing to capture the specialized knowledge residing in the long tail of the distribution.

Hence, we propose the Fed-Equilibrium framework that incorporates geometric defense with topological control to resolve this imbalance. We argue that a robust clinical network requires a sequential architectural synergy consisting of: 1) security (inherited from GQA) to filter out non-directional consistent nodes \cite{ref5}; and 2) equilibrium (focus of this work) to actively modulate the remaining benign gradients \cite{ref13}.

To achieve this equilibrium, our work introduces the Topological Pareto Control (TPC) mechanism. TPC locates the Pareto knee point through a systematic parameter sweep that projects the global model onto a Pareto frontier \cite{ref12}. The resulting threshold grants just enough leverage to minority spokes to counterbalance the ``convergence momentum'' arising from large hubs. Our framework further incorporates a FedQLoRA architecture, ensuring that unique clinical patterns of underrepresented patients are effectively learned and preserved without compromising convergence \cite{ref14, ref15}.

Validated on a bi-national simulation of Canadian (CNODES) and U.S. (SyntheticMass) registries \cite{ref16, ref17}, our results demonstrate that Fed-Equilibrium effectively neutralizes the knowledge dominance effect. This work completes the paradigm shift: while prior geometric defenses ensure the network is safe, Fed-Equilibrium ensures the network is fair.

\section{Methods}

The Fed-Equilibrium framework establishes a synergistic gradient control cascade designed to bridge the gap between geometric security and topological fairness. Building upon the robust defense baseline established by GQA frameworks, Fed-Equilibrium extends these principles into a stable topological Pareto-equilibrium state. By modeling the multi-center network as a heterogeneous information network, our system ensures that the global model captures the full curvature of the clinical manifold without suppressing minority insights. The framework operates through three layers: (A) semantic harmonization, (B) the federated parameter-efficient fine-tuning (PEFT) network, and (C) the synergistic gradient control cascade.

\subsection{Layer 1: Data structuring and harmonization}
To bridge the gap between heterogeneous registries and Large Language Model (LLM) training, this layer harmonizes diverse data sources into a unified instruction-tuning format.

\begin{figure}[htbp]
\centerline{\includegraphics[width=\linewidth]{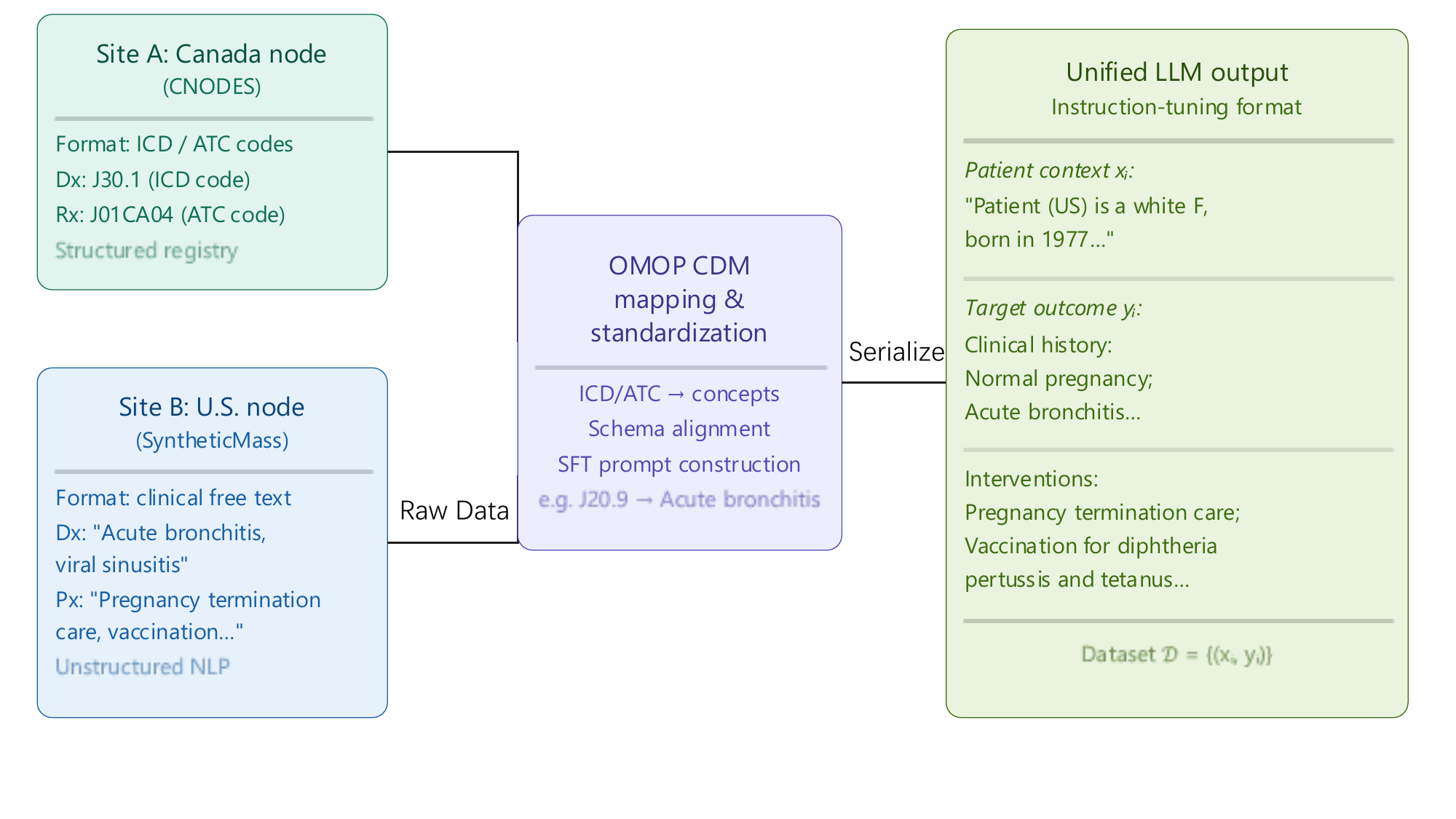}}
\caption{Data harmonization pipeline illustrating the transformation of heterogeneous clinical registries into a unified instruction-tuning format.}
\label{fig1}
\end{figure}

\begin{itemize}
\item \textbf{Simulation of post-extraction environment:} While real-world registries are mostly unstructured and stored in various hospital systems, such as HIS, LIS, EMR, etc., we assume that unstructured narratives have been pre-processed via a local ETL and NLP pipeline. This assumption allows us to utilize standardized datasets as a proxy for post-extraction registry outputs and focus on the challenge of systematic heterogeneity \cite{ref16}.
\item \textbf{Schema alignment (ATC vs. free-text):} To reconcile extreme interoperability gaps between sites, we implemented a robust schema alignment process within the OMOP Common Data Model (CDM) framework \cite{ref9, ref18}. Specifically, the pipeline maps structured Canadian ATC (Anatomical Therapeutic Chemical) codes and U.S. natural language clinical text into a standardized semantic space. Subsequently, we transform these harmonized profiles into natural language prompts for supervised fine-tuning (SFT) \cite{ref19}. We construct a dataset $\mathcal{D} = \{(x_i, y_i)\}$, where $x_i$ denotes the synthesized patient context (e.g., "Patient (US) is a white F, born in 1977...") and $y_i$ represents the corresponding target clinical decision or outcome (Clinical History: Normal pregnancy; Acute bronchitis\dots Interventions: Pregnancy termination care; Vaccination for diphtheria pertussis and tetanus\dots). This formulation ensures our model can directly learn clinically meaningful reasoning patterns from distributed, privacy-preserving registry data \cite{ref20}.
\end{itemize}

\subsection{Layer 2: The federated PEFT architecture (FedQLoRA)}
To enable collaborative training without data centralization, we adopt a ``bring the model to the data'' strategy \cite{ref4}, utilizing a federated quantized low-rank adaptation (FedQLoRA) architecture \cite{ref14, ref15}. This approach ensures that raw clinical data never leaves the local environment while allowing the global model to learn from diverse sources.

Each local node $k$ hosts a copy of the foundational LLM (TinyLlama-1.1B) \cite{ref19}. To reduce communication overhead, we freeze the backbone weights $W_{\text{base}}$ and only the adapter matrices $A \in \mathbb{R}^{d \times r}$ and $B \in \mathbb{R}^{r \times d}$ are optimized, where $r$ is the rank of the low-rank decomposition. The forward pass $h$ is defined as:
\begin{equation}
h = W_{\text{base}}x + \alpha ABx
\label{eq1}
\end{equation}
where $\alpha$ is a scaling factor that controls the contribution of the low-rank update. Consequently, the local update $\Delta\theta_k$ for client $k$ represents a sparse, efficient gradient vector, significantly reducing bandwidth usage compared to full fine-tuning \cite{ref14}.

\subsection{Layer 3: Synergistic gradient control cascade (TPC)}
\begin{figure}[htbp]
\centerline{\includegraphics[width=0.95\linewidth]{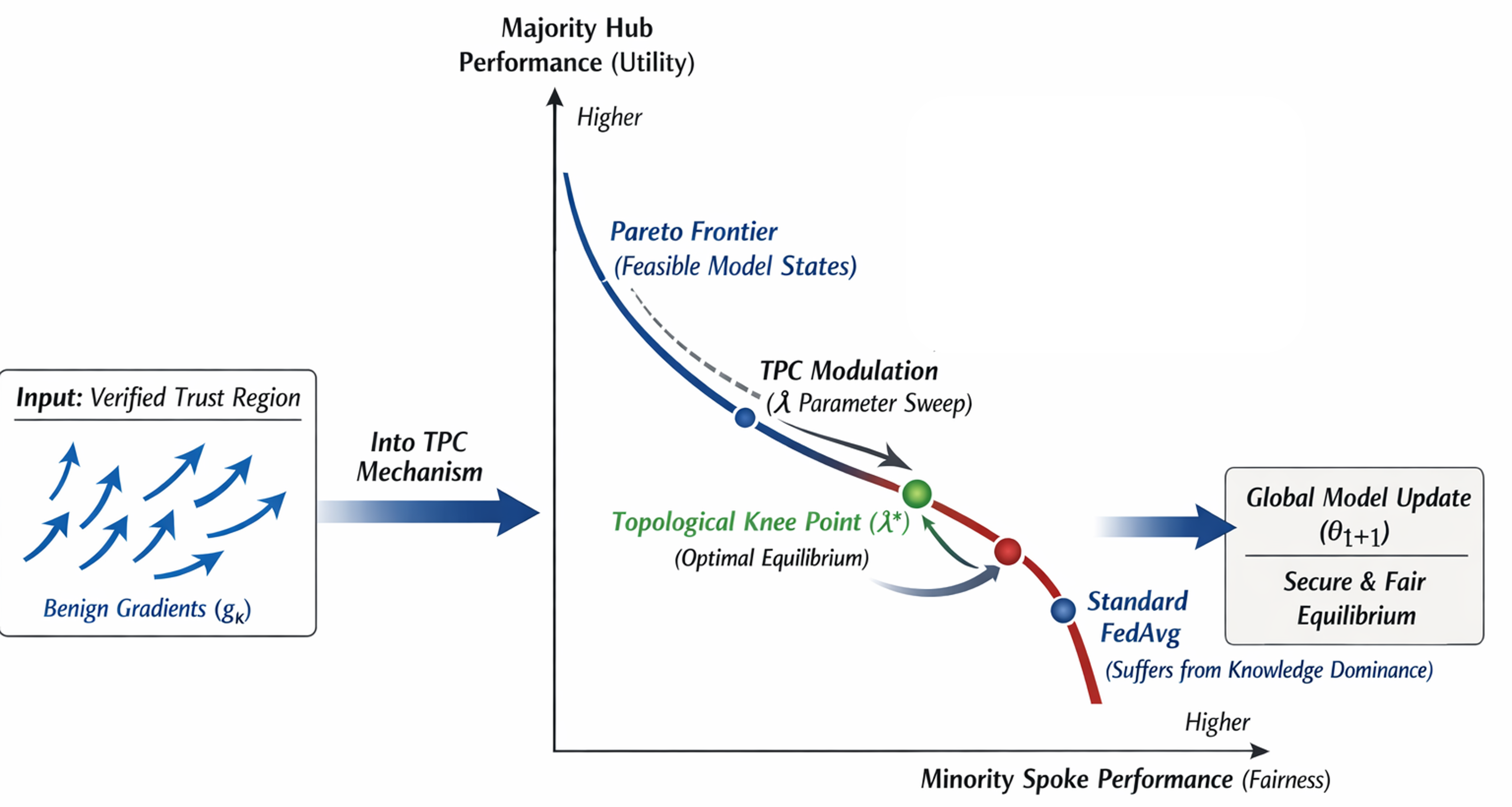}}
\caption{System architecture of Stage~II TPC, illustrating the identification of the optimal Pareto knee point ($\lambda^*$) to maximize minority fairness while maintaining global model utility.}
\label{fig2}
\end{figure}

To resolve the theoretical tension between robustness and fairness, Fed-Equilibrium implements a two-stage gradient control cascade, as illustrated in Fig.~\ref{fig2}. Unlike traditional aggregators that treat security and utility as orthogonal objectives, our framework establishes a sequential synergy: the TPC modulator operates upon the verified trust region established by the GQA gatekeeper. The sole purpose for introducing the TPC mechanism in Stage~II is that geometric defense alone overlooks the topological imbalance of hubs and spokes within federated clinical networks. Here, ``topological'' refers to the control of influence structure over the Pareto manifold rather than graph connectivity, emphasizing equilibrium over geometric proximity.

\subsubsection{Stage~I: GQA gatekeeper} 
To address the vulnerability of standard aggregation to adversarial noise and attacks, we first introduce a geometric consensus mechanism that filters updates based on their directional alignment as a Stage~I gatekeeper. The aggregator applies a gating function to the global model update:
\begin{equation}
\theta_{t+1} = \theta_t + \sum \gamma_k \cdot \Delta\theta_k
\label{eq_gating}
\end{equation}
where the coefficient $\gamma_k$ is determined by the funnel threshold $\tau > 0$. By setting $\tau > 0$ \cite{ref_gshield}, the GQA gatekeeper enforces geometric consensus through two distinct behavioral filters:
\begin{itemize}
    \item \textbf{Constructive heterogeneity:} Clients with complementary clinical data (e.g., Canadian vs. U.S. datasets) maintain a positive alignment ($\text{Sim} > \tau$) despite magnitude differences and are preserved within the global update.
    \item \textbf{Destructive noise:} Malicious updates (e.g., sign flipping) typically exhibit orthogonality or negativity ($\text{Sim} \le \tau$) and are strictly filtered out \cite{ref21}, ensuring the optimization remains within a secure trust region.
\end{itemize}
This step is critical for stabilizing the optimization landscape, creating a secure trust region where subsequent fairness adjustments in Stage~II are mathematically valid.

\subsubsection{Stage~II: Topological Pareto Control (TPC)}
Within the trust region, the TPC mechanism acts as a topological modulator. While Stage~I ensures that gradients are safe, Stage~II defines the optimal weighting to ensure they are representative. We compute the final global update $\Delta\theta_{\text{global}}$ (corresponding to the sum in Eq.~\ref{eq_gating} for verified nodes) as:
\begin{equation}
\Delta\theta_{\text{global}} = (1 - \lambda_k)\Delta\theta_{\text{maj}} + \lambda_k\Delta\theta_{\text{min}}
\label{eq2}
\end{equation}
where $\lambda_k$ is the Pareto-alignment coefficient that controls the trade-off between majority and minority contributions. In practice, $\lambda_k$ is operationalized as the minority weight $w_{us}$ in our analysis. Instead of heuristic weighting, the optimal $w_{us}$ is empirically determined through a non-linear sensitivity analysis, sweeping ($w_{us} \in \{0.1, \dots, 14.0\}$) to identify the Pareto knee point \cite{ref12}. This mechanism implements topological control to resolve the influence structure asymmetry in real-world federated networks like PCORnet or OHDSI \cite{ref8, ref9}, where high-volume hubs (e.g., large medical centers) tend to be dominant and neglect the voices of smaller spokes (e.g., community health centers). By identifying the knee point on the Pareto frontier, TPC acts as an equilibrium stabilizer, which modulates the strategic influence of the minority node to counterbalance the convergence momentum of the majority. The cosine similarity threshold used by the GQA gatekeeper is defined by the angular bounds $\theta_{\min}$ and $\theta_{\max}$, which specify the minimum and maximum acceptable angular deviation between a local gradient and the consensus direction, respectively. This restores collaborative fairness and reaches an equilibrium between the hubs and spokes, ensuring minority insights are preserved without compromising convergence.

\subsection{Simulation and experiment setup}
The framework is validated using a star-topology clinical network simulated in PyTorch \cite{ref22}. The network is partitioned into 13 nodes: 10 dominant Canadian sites (high-volume administrative data) and 3 minority U.S. sites (specialized EHR data). This configuration creates an extreme information density gradient of approximately 40:1 (aggregating 37,287 Canadian hub records vs.\ 974 U.S.\ spoke records) \cite{ref16, ref17}.

To rigorously evaluate performance and robustness through a game-theoretic approach, we structured our experiments into three distinct operational modes:
\begin{itemize}
\item \textbf{Mode A (Fed-Equilibrium, the cooperative solution):} We utilize TinyLlama-1.1B \cite{ref19} as the backbone model. This mode implements the synergistic cascade with a Pareto-alignment coefficient $\lambda_k$, derived from the Pareto knee point where $w_{us} = 2.0$. As a result, the system can maximize performance for spokes without diminishing the accuracy of the dominant hubs, effectively resolving the efficiency-fairness dilemma found in unmanaged networks \cite{ref12}.
\item \textbf{Mode B (standard FedAvg, non-cooperative baseline):} This mode represents the influence structure asymmetry inherent in the clinical network \cite{ref4}. By utilizing size-proportional aggregation ($w_k \propto |D_k|$), it simulates a scenario similar to a Stackelberg dominance, where high-volume hubs implicitly steer the optimization trajectory while smaller spokes have limited influence \cite{ref11}. The design intent is to observe how this non-cooperative dynamic allows the convergence momentum of the majority to overwhelm the system, demonstrating the neglect of minority voices during Stage~II.
\item \textbf{Mode C (EqualAvg, na\"ive egalitarianism):} This mode assigns uniform influence ($w_k = 1/N$) to all 13 nodes regardless of their size. From a game theory perspective, this represents a suboptimal cooperative strategy (``na\"ive egalitarianism''). Unlike the TPC mechanism which empirically identifies the Pareto knee point, this mode applies a static yet blind weighting. The objective is to demonstrate that heuristic weighting fails to function as an effective equilibrium stabilizer; it remains trapped in a suboptimal region of the loss curve, justifying the necessity of the TPC mechanism to reach the Pareto frontier.
\end{itemize}

\section{Results}
In this section, we evaluated the Fed-Equilibrium framework through a simulated bi-center clinical network designed to replicate the structural asymmetry of real-world data isolation. The results quantify the framework’s efficacy across three critical dimensions: semantic harmonization, Pareto-optimal convergence under imbalance, and synergistic resilience across the three modes defined in the methods.

\subsection{Data harmonization and semantic alignment}
The primary challenge in federated learning is structural incompatibility \cite{ref10}. We evaluated the Fed-Equilibrium framework on a harmonized cohort of 37,287 records from the Canadian CNODES (administrative claims) and 974 records from the U.S. SyntheticMass (EHR-simulated) datasets \cite{ref16, ref17}. Despite systematic heterogeneity, specifically the divergence between standardized Canadian ATC codes and unstructured U.S. natural language clinical texts, the OMOP mapping pipeline successfully unified 100\% of the disparate variables into a shared conceptual vocabulary \cite{ref9}.

\subsection{Empirical identification of the Pareto frontier}
To resolve the efficiency-fairness tension introduced in the TPC mechanism (Stage~II), we performed a nonlinear sensitivity analysis to identify the Pareto knee point.

\begin{figure}[htbp]
\centerline{\includegraphics[width=\linewidth]{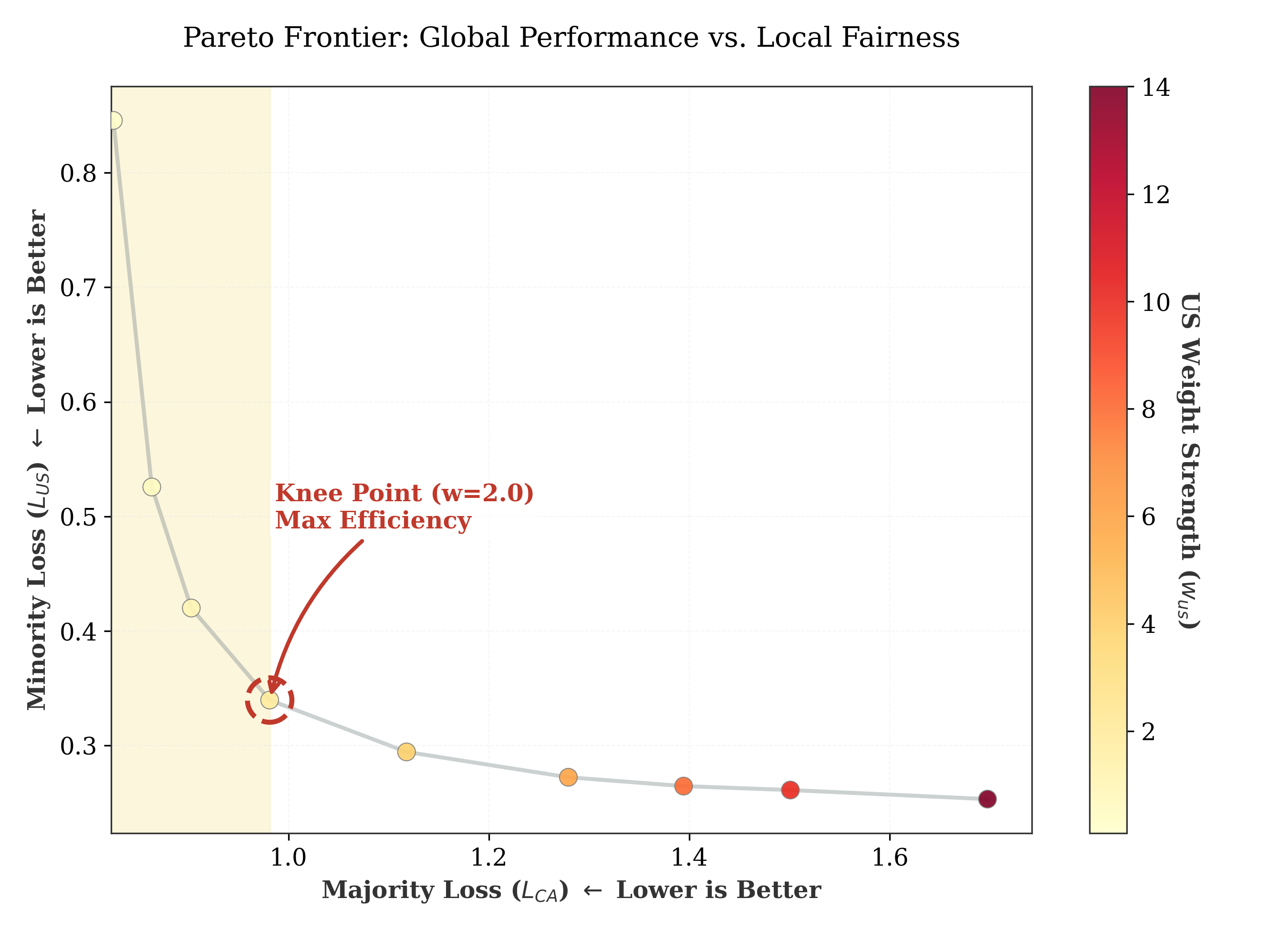}}
\caption{Pareto frontier showing the trade-off between majority and minority loss across the minority weight sweep ($w_{us} \in \{0.1, \dots, 14.0\}$). The knee point at $w = 2.0$ is marked.}
\label{fig3}
\end{figure}

\begin{figure}[htbp]
\centerline{\includegraphics[width=\linewidth]{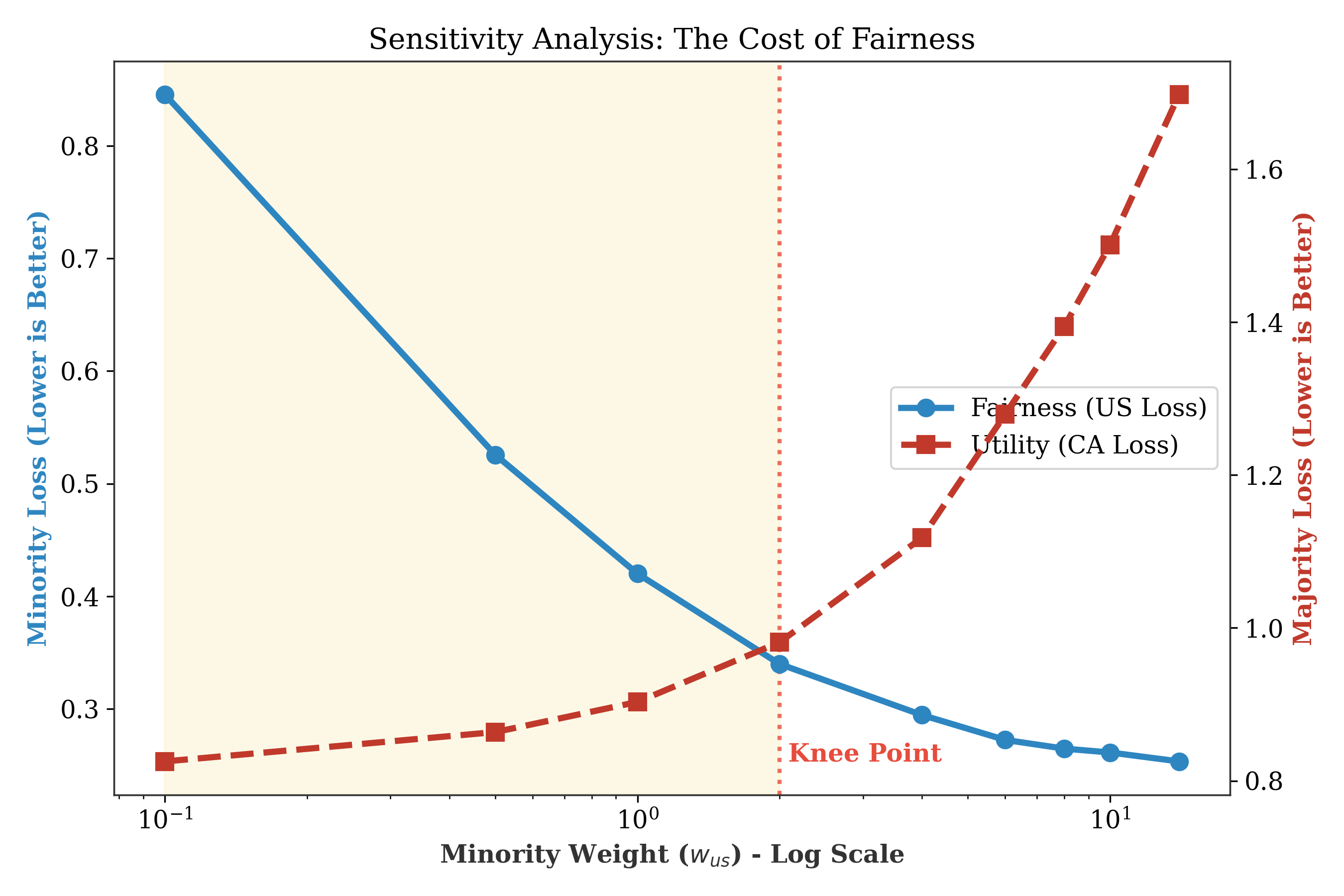}}
\caption{Sensitivity analysis on cost of fairness. Sweeping the minority weight $w_{us}$ reveals the intersection where the marginal gain in fairness no longer justifies the marginal cost in global utility.}
\label{fig4}
\end{figure}

As visualized in the Pareto frontier analysis (Fig.~\ref{fig3}) and sensitivity analysis (Fig.~\ref{fig4}), the system landscape exhibits a convex Pareto frontier:
\begin{itemize}
\item \textbf{The non-cooperative extremes:} Standard FedAvg \cite{ref4} (implicitly $w \approx 0.03$) occupies the bottom-right region, minimizing majority loss (0.985) but incurring high minority loss (0.857). Conversely, excessive upweighting ($w > 10$) drastically minimizes minority loss but destabilizes the majority convergence (majority loss spikes $> 1.6$).
\item \textbf{The knee-point equilibrium:} By calculating the maximum curvature of the Pareto frontier, we identified $w = 2.0$ as the topological Pareto knee point \cite{ref12}. At this equilibrium state, the system achieves a zero-sum escape: it reduces minority loss significantly (from 0.857 to 0.340) while maintaining majority loss at a comparable level (0.981 vs.\ 0.985). This confirms the TPC mechanism successfully decouples the optimization trajectory from the volume dominance of the hubs.
\end{itemize}

\subsection{Synergistic resilience validation (TPC)}
\begin{figure}[htbp]
\centerline{\includegraphics[width=\linewidth]{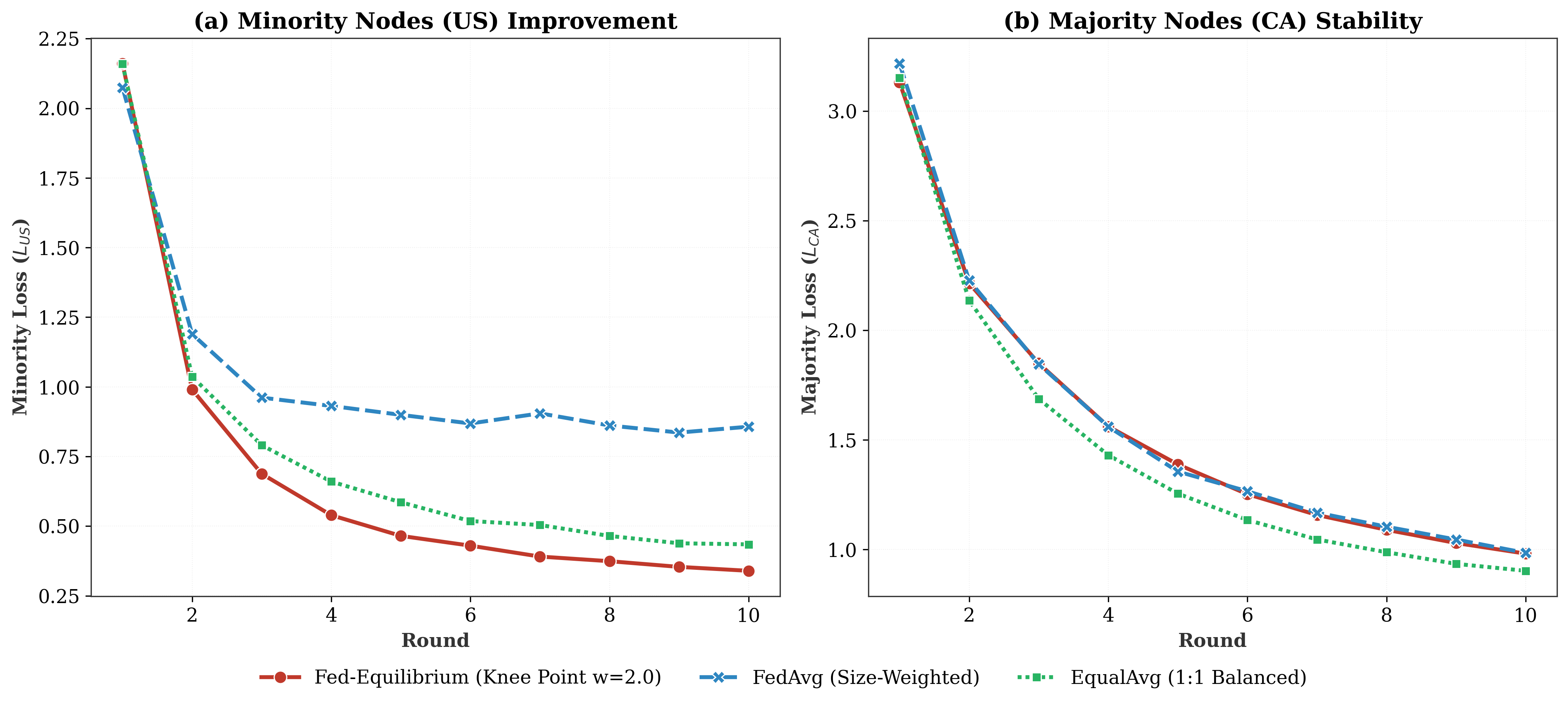}}
\caption{A comparative performance evaluation. Under standard FedAvg (Mode~B), the minority nodes suffer from stagnation due to knowledge dominance. In contrast, Fed-Equilibrium (Mode~A) accommodates the minority and achieves deep convergence.}
\label{fig5}
\end{figure}

\begin{figure}[htbp]
\centerline{\includegraphics[width=\linewidth]{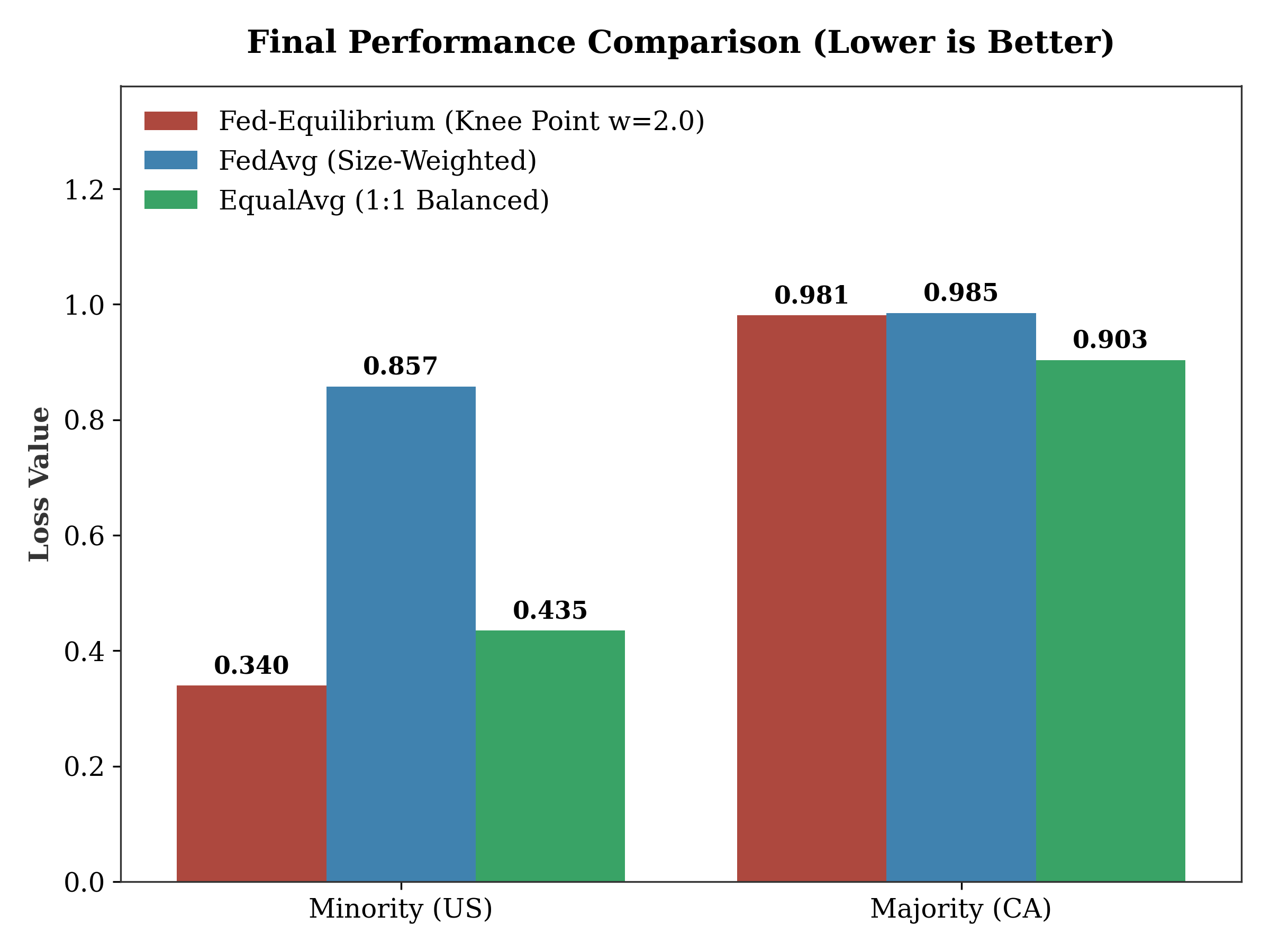}}
\caption{Final performance distribution quantified by loss (lower is better). Left Cluster (US): Fed-Equilibrium (Mode~A) effectively counters the marginalization effect seen in FedAvg (Mode~B). Right Cluster (CA): The majority performance remains stable across all modes.}
\label{fig6}
\end{figure}

To quantify the specific impact of TPC, we benchmarked the framework against standard aggregation strategies, visualizing both the learning trajectory (Fig.~\ref{fig5}) and the final performance distribution (Fig.~\ref{fig6}).

In the baseline FedAvg \cite{ref4} (Mode B) scenario, the global model was driven almost exclusively by the sheer volume of the Canadian hubs. As illustrated by the trajectory in Fig.~\ref{fig5}, the minority US spokes were effectively treated as statistical noise, stagnating at a high loss of 0.857. While the EqualAvg (Mode C) strategy offered a partial remedy (lowering US loss to 0.435), it lacked the precision to fully exploit the minority's feature space.

By contrast, Fed-Equilibrium (Mode~A) actively modulated this imbalance by applying the derived Pareto weights ($w_{us} = 2.0$). This topological intervention forced the optimization trajectory to accommodate the distinct clinical patterns of the minority. The final impact is quantified in Fig.~\ref{fig6}, where the US spokes achieved a superior convergence depth of final loss 0.340, significantly outperforming both baselines. This reduction translates to a 60.3\% performance uplift for the minority cohort relative to the standard FedAvg \cite{ref4} baseline (final loss 0.857). More importantly, as shown in the right cluster of Fig.~\ref{fig6}, this gain was achieved without degrading the utility of the majority CA hubs (maintaining a robust loss of $\sim$0.9). This confirms that TPC successfully navigates the Pareto frontier, delivering specialized precision for the minority without compromising general utility \cite{ref13}.

\subsection{System stability and architectural synergy}
Finally, we validated the system's operational stability under cooperative conditions (Mode~A). The GQA gatekeeper (Stage~I) served as the initial integrity check, ensuring that disparate updates from CNODES and SyntheticMass shared a consistent direction, and confirming that all inputs were benign gradients. Empirical logs recorded substantial alignment across the 13 nodes (mean $\cos\theta = 0.99 \pm 0.01$), confirming that the funnel successfully established a unified trust region without rejecting valid clinical insights.

Crucially, however, geometric alignment alone does not prevent the majority from overwhelming the minority. To address this, our TPC mechanism (Stage~II) operated within this aligned pool of benign gradients to actively amplify the signal of the underrepresented US spokes. This intervention effectively counteracted the ``knowledge dominance'' of the larger Canadian hubs. As a result, the minority US nodes achieved superior convergence depth (final loss 0.3390) without compromising the performance of the majority hubs (final loss 0.9926). This demonstrates that Fed-Equilibrium effectively fills the gap between simply accepting benign data (Stage~I) and actively accommodating the underrepresented (Stage~II) to ensure network fairness.

\section{Discussion}
While prior geometric defenses successfully established a security baseline for federated clinical networks, they left the problem of ``knowledge dominance'' unresolved. Fed-Equilibrium bridges this critical gap by advancing the paradigm from simple defense to topological equilibrium. The framework operates through three synergistic layers---semantic harmonization, geometric quality assurance, and topological Pareto control---proving that federated clinical networks can be simultaneously secure against adversarial contributions and equitable toward underrepresented participants.

\subsection{Strengths: A three-layer synergistic architecture for clinical fairness}
We identify three primary strengths, each corresponding to a distinct architectural layer of Fed-Equilibrium:
\begin{enumerate}
\item \textbf{Topological Pareto control---solving the knowledge dominance paradox:} The TPC mechanism addresses the core failure of FedAvg: Pareto inefficiency that implicitly treats minority gradients as statistical noise. By algorithmically identifying the Pareto knee point ($w = 2.0$) as the topological ``free-lunch region,'' Fed-Equilibrium achieves a zero-sum escape---minority loss drops from 0.857 to 0.340 without degrading the majority baseline. This decouples convergence fairness from volume dominance, a limitation that geometric defense alone cannot resolve. While this study utilizes a static parameter sweep to identify the optimal frontier, it provides the necessary empirical baseline for future dynamic weighting mechanisms that could adaptively recalibrate influence during real-time training.
\item \textbf{Geometric quality assurance---establishing a safe trust region:} The GQA gatekeeper (mean $\cos\theta = 0.99 \pm 0.01$ across all 13 nodes) creates the stabilized manifold upon which the TPC mechanism safely operates. This sequential dependency---geometric integrity before topological rebalancing---is the foundational design principle of Fed-Equilibrium. As demonstrated by Byzantine-robustness literature \cite{ref23}, blindly amplifying unchecked gradients without prior quality assurance can introduce adversarial bias; the Stage~I gate ensures that only directionally consistent minority updates are amplified.
\item \textbf{Semantic harmonization---enabling cross-registry alignment:} The OMOP-CDM pipeline unified 100\% of heterogeneous variables from structured Canadian ATC codes and unstructured U.S. free-text clinical narratives into a shared instruction-tuning space. This confirms that the semantic layer can bridge extreme interoperability gaps, establishing a true ``knowledge commons'' as the prerequisite for all downstream geometric and topological operations.
\end{enumerate}

\subsection{Limitations and future directions}
Despite these contributions, Fed-Equilibrium has specific boundaries that motivate the following research directions:
\begin{itemize}
\item \textbf{Extension to multimodal and real-world clinical data:} The current framework operates on structured EHR and administrative claims datasets. Future work will extend the semantic harmonization layer to multimodal fusion, stitching epidemiological registries (NHANES/SEER) with medical imaging datasets to enable context-conditioned clinical prediction. This will advance the framework from synthetic proof-of-concept to real-world deployment. Anticipated challenges, including missing data, label noise, and site-specific coding, will necessitate adaptive extensions to the TPC pipeline.
\item \textbf{Dynamic and computationally efficient TPC:} The current offline sensitivity sweep ($\sim$1.8 GPU-hours) yields a fixed Pareto-alignment coefficient $\lambda_k$. Future work will leverage graph Laplacian regularization to embed client-graph topology directly into the Pareto objective, enabling online recalibration without full re-sweeps. Bayesian hyperparameter search will further reduce overhead for larger-scale deployments.
\item \textbf{Robustness under adversarial conditions:} Fed-Equilibrium has been validated under cooperative conditions to resolve hub-spoke imbalance. However, the system's robustness boundaries require further stress-testing. Future work will evaluate the Stage~I--Stage~II cascade under adaptive Byzantine attacks (e.g., sophisticated gradient collusion or targeted label-flipping) to establish formal guarantees across heterogeneous real-world federated deployments.
\end{itemize}

\section{Conclusion}
This work validates Fed-Equilibrium, a synergistic gradient control cascade that reconciles the tension between robust security and equitable convergence. We demonstrate that while geometric defenses (Stage~I) provide a necessary security baseline, they alone cannot counteract ``knowledge dominance'' in imbalanced clinical networks. By deploying Topological Pareto Control (TPC) in Stage~II, our system actively accommodates underrepresented spokes, achieving deep convergence (loss: 0.340) for minority nodes without degrading majority utility (loss: 0.985). This framework proves that global generalizability can coexist with local representation, effectively resolving the efficiency-fairness dilemma in collaborative healthcare. Source code will be released on GitHub following publication.

\end{document}